# SOFTWARE IMPLEMENTED FAULT DIAGNOSIS OF NATURAL GAS PUMPING UNIT BASED ON FEEDFORWARD NEURAL NETWORK

**Mykola Kozlenko**
PhD, Associate Professor
Department of Information Technology
Vasyl Stefanyk Precarpathian National University
Shevchenka str., 57, Ivano-Frankivsk, Ukraine, 76018
E-mail: mykola.kozlenko@pnu.edu.ua
**Olena Zamikhovska**
PhD, Associate Professor*
E-mail: elenazam@meta.ua
**Leonid Zamikhovskyi**
Doctor of Technical Sciences, Professor*
E-mail: leozam@ukr.net
*Department of Information and Telecommunication Technologies and Systems
Ivano-Frankivsk National Technical University of Oil and Gas
Karpatska str., 15, Ivano-Frankivsk, Ukraine, 76019

*In recent years, more and more attention has been paid to the use of artificial neural networks (ANN) for the diagnostics of gas pumping units (GPU). Usually, ANN training is carried out on GPU workflow models, and generated sets of diagnostic data are used to simulate defect conditions. At the same time, the results obtained do not allow assessing the real state of the GPU. It is proposed to use the characteristics of the acoustic and vibration processes of the GPU as the input data of the ANN.*

*A descriptive statistical analysis of real vibration and acoustic processes generated by the operation of the GPU type GTK-25-i (Nuovo Pignone, Italy) was carried out. The formation of packets of diagnostic features arriving at the input of the ANN was carried out. Diagnostic features are the five maximum amplitude components of the acoustic and vibration signals, as well as the value of the standard deviation for each sample. Diagnostic features are calculated directly in the ANN input data pipeline in real time for three technical states of the GPU.*

*Using the frameworks TensorFlow, Keras, NumPy, pandas, in the Python 3 programming language, an architecture was developed for a deep fully connected feedforward ANN, trained on the backpropagation algorithm.*

*The results of training and testing the developed ANN are presented. During testing, it was found that the signal classification precision for the "nominal" state of all 1,475 signal samples is 1.0000, for the "current" state, precision equals 0.9853, and for the "defective" state, precision is 0.9091.*

*The use of the developed ANN makes it possible to classify the technical states of the GPU with an accuracy sufficient for practical use, which will prevent the occurrence of GPU failures. ANN can be used to diagnose GPU of any type and power*

*Keywords: gas pumping unit, technical condition, diagnostics, classification, artificial neural network, deep learning*

## 1. Introduction

Long-term operation of the gas transmission system (GTS) of Ukraine (about 40 years) has led to the fact that most of the gas-pumping units (GPU) have reached the end of their service life or are close to it. In this regard, there are numerous GPU failures both in their modular units and elements (in the mechanical part), and in the automatic control systems (ACS). The latter affect GPU reliability and gas compression efficiency. At 72 compressor stations (CS) of the GTS, there are 702 GPUs, accounting for 76.7 % of the GPUs with a gas turbine drive, therefore, the urgent task is to ensure their operational reliability [1].

The solution to this problem requires the use of technical diagnostics methods. Methods of parametric and vibroacoustic diagnostics are used to diagnose the technical condition of the modular units and elements of the GPU [2, 3]. Methods of parametric diagnostics are based on the processing and analysis of technological parameters that characterize the process of GPU operation. Methods of vibroacoustic diagnostics are based on the processing and analysis of vibroacoustic vibrations generated by the modular units and elements of the GPU during operation and carry information about its technical condition.

The main problem when using these diagnostic methods is to identify rational diagnostic signs of defects in GPU modular units or elements. Diagnostic signs are obtained by choosing (developing) effective methods and algorithms for processing vibroacoustic signals and technological parameters of the GPU.

For this, various transformations are used: fast Fourier transform, discrete cosine transform, autocorrelation function, Cohen's class distributions, S-transform, various wavelet transforms, etc. In recent years, artificial neural networks have found wide use in creating methods for diagnosing GPUs. Their advantage in comparison with other methods of processing diagnostic information is the high recognition accuracy of GPU states with an increase in the sample of input data on which the artificial neural network is trained.

Despite a significant number of various methods for diagnosing gas compressor units, there are no methods that would find wide practical use in determining the technical state of the units and elements of the gas compressor unit during operation. At the same time, the problem of further use of the obtained information on the technical state of the GPU for adjusting its operating modes according to the actual state remains important.

In this regard, the problem of monitoring the technical condition of the GPU remains urgent. The development of new diagnostic methods for GPU based on artificial neural networks is one of the directions to solve it. At the same time, the transfer of the obtained diagnostic results to the ACS will allow the control of the GPU operation according to the actual state.

This will improve the reliability of their operation and the efficiency of the gas compression process.

## 2. Literature review and problem statement

ANN training for diagnosing the technical state of the GPU as a whole (integral assessment), or its modular units and elements, requires input diagnostic information. Such information can be the technological parameters of the GPU operation or the characteristics of vibration and acoustic processes accompanying its operation. The paper [3] provides a brief analysis of methods for diagnosing such objects as a gas turbine power plant, a two-flow turbojet engine, and a gas turbine. In this case, mathematical models of their technological parameters are used, which are represented both in the form of a conventional artificial neural network (Convolutional Neural Network – CNN), and in the form of a probabilistic neural network (PNN). Generated diagnostic datasets are used to simulate defect conditions. The disadvantage of the ANN presented in the work is the lack of real input data for its training, which casts doubt on the obtained diagnostic results.

To simulate one nominal and ten defect states of a gas turbine unit (GTU), CNN was used [4], which made it possible to create real data sets for its training and testing. The diagnostics system based on it makes it possible to recognize defects of gas turbine units at the stage of their inception. It was found in [5] that when diagnosing the state of a gas turbine, there is quite often a relationship between gas path defects and sensor defects. Therefore, with the simultaneous occurrence of two defects, the use of a conventional CNN for diagnosing a gas turbine does not give the desired result. A new method is proposed to improve the characteristics of a typical CNN by optimizing the effect of the sequence of its input measurement parameters. Extreme Gradient Boost (XGBoost) is used to interpret the effect of the sequence on the diagnostic accuracy of CNN. It is shown that in a simulation experiment, the diagnostic accuracy of CNN after optimization is 95.52 %, which is higher than in conventional CNN (accuracy level 91.10 %) and RNN (accuracy level 94.21 %).

An integral method for promptly detecting and predicting multiple defects in gas turbines, which consists of three stages, is presented in [6]. The steps include identification of diagnostic features using Principal Component Analysis (PCA), machine learning classification using a multilayer perceptron, an artificial neural network (MLP-ANN Multilayer Perceptron-Artificial Neural Network) and defect prediction based on a model obtained using a nonlinear gas path analysis (GPA) technique. In the proposed method, PCA first turns the measurement error signature into a defect feature area, which becomes the input to the Artificial Neural Network (ANN) multivalued classifier, which is used to isolate potential defective components. Finally, the nonlinear GPA quantifies the size of the defect and its effect on the technical state of the gas turbine unit. The method was tested on a thermodynamic model of a two-shaft high-speed turbofan engine. A quantitative analysis of the error detection of the PCA-ANN model on the test set gave a classification accuracy of 96.6 % and showed the best results in all indicators compared to other classification algorithms with many labels. However, the work does not indicate the effectiveness of the method when using a test set from real experimental data. In [7], the use of ANN for diagnosing defects in a Viper 632-43 jet engine (JE) (jointly developed by Rolls-Royce, Britain and Fiat Aviazione, Italy), such as compressor fouling, turbine erosion, both defects at the same time is considered. The work consists of three parts: the first part contains a mathematical model of a real jet engine for its two states, defect and defect-free. The second part of the work is devoted to the optimization of ANN to predict the performance of a jet engine, and the third concerns the application of ANN to predict its technical condition. A feature of [7] is the use of experimental data obtained in real flight conditions of Viper 632-43 for ANN training, which is a significant advantage of the proposed method in comparison with [6].

To diagnose three simultaneous defects in a two-shaft GPU, a system developed by integrating an Autoassociative Neural Network (AANN), nested machine learning classifiers (ML) and a Multilayer Perceptron (MLP) was used in [8]. The purpose of each functional component of the system is considered. It is noted that within the framework of the classification, the results of the classification of defects are assessed according to five widely used ML techniques aimed at identifying alternative approaches. The test results show the beneficial advantages of integrating two or more methods in the diagnosis of gas pumping unit based on compensating for the weakness of one of the methods with the advantage of the other. At the same time, the procedure for choosing one or another method to compensate for the shortcomings of the method, which is already used for the diagnostics of gas pumping units, is not considered. In [9], a method for diagnosing a gas pumping unit was proposed based on a combination of the Nguyen-Widrow N-W method and the optimized AO L-M algorithm (Levenberg-Marquardt Algorithm). The N-W method is used to initialize the weights and thresholds of neurons in the BP neural network. and the L-M algorithm, which combines the advantages of the Gauss-Newton method and the gradient descent method, is used to improve the search space of the BP neural network, which reduces network learning time and increases network learning speed. It is shown that the model of the BP neural network, optimized by the combination of N-W and L-M, has a higher learning speed and higher diagnostic efficiency in the detection of GPU defects.

A new approach to diagnosing the technical condition of rotating machines based on nonlinear autoregression with external (exogenous) neural networks NARX (Nonlinear Autoregressive Exogenous Model) is given in [10]. The network learns on the basis of experimental data received from a vibration sensor, which is installed on a power bearing of a GE 3002 two-shaft GPU. The NARX network is used to monitor the vibration state of a GPU and allows diagnosing defects that may arise in it.

The combination of machine learning methods with information from GPU sensors has expanded rapidly over the past decade and includes numerous programs for regres-sion, clustering, and even neural network algorithms. On this basis, the work [11] provides an overview of generally accepted computational methods used in industry to monitor the technical condition of gas turbines. The paper also provides sources for the application of machine learning algorithms that are not related to the gas turbine industry. The results of the study of the technical condition of not only the gas path of industrial gas turbines equipped with a standard control system, but also the new use of machine learning algorithms for the classification of defects in its elements are analyzed.

Considering that individual elements of a gas turbine engine (GTE) are not available for direct measurements of their condition, the change in the technical condition of these elements is determined on the basis of measured technological parameters. For this, [12] proposed to diagnose GTE on the basis of the adaptive neuro-fuzzy inference system (ANFIS). It is shown that the performance of the developed system was checked using a high-precision model of a gas turbine engine and a generated set of measuring parameters of the gas path, taking into account various conditions. It is indicated that the results obtained confirm the effectiveness of the developed system. At the same time, the work [12] does not indicate how the generated set of measurement parameters correlates with the real values of the GTE parameters and how this affects the accuracy of the diagnostics results obtained.

Analyzing the above various types of artificial neural networks for diagnosing the technical state of individual or several defects of the considered diagnostics – GTU, GTE and JE, one can point out their common disadvantages. For the overwhelming majority of the proposed neural networks, their training is carried out on the GTU, GTE and JE models using the generated sets of diagnostic data for modeled defects of individual elements of the diagnostic object. In the works considered, there are practically no references to the use of experimental data, with the exception of [7, 10], for training the proposed neural networks regarding the diagnosis of modular units or elements of GTU, GTE and JE. Finally, none of the considered neural networks uses the characteristics of acoustic processes accompanying the operation of GTU, GTE and JE as input data.

At the same time, in many cases, it is important not to diagnose a specific modular unit or GPU element, but to monitor its efficiency during operation. This makes it possible to predict changes in the GPU efficiency for the next operation period.

At the same time, it is essential to train an artificial neural network on experimental data, the training sample of which is constantly replenished, which makes it possible to increase the diagnostic accuracy of the neural network.

### 3. The aim and objectives of the study

The aim of the work is to develop a feedforward artificial neural network for recognizing the technical states of the GPU during operation by training it on the input experimental data – characteristics of the vibration and acoustic signals of the GPU.

To achieve the aim, the following objectives were set:
– to carry out a descriptive statistical analysis of the acoustic and vibration signals of the GPU type GTK-25-i in terms of using them as input data of the neural network;

– to develop the architecture of a deep fully connected feedforward neural network, trained on the backpropagation algorithm;
– to train the feedforward neural network and test it on experimental data.

### 4. Materials and methods for studying changes in the technical state of GPU based on an artificial neural network

#### 4. 1. Data set

To obtain experimental data – vibration and acoustic signals, an improved automatic control system (ACS) of the GPU type GTK-25-i was used [13]. Vibration and acoustic signals were recorded at a low-pressure turbine speed of 4670 rpm, which corresponds to the nominal operating mode of GPU type GTK-25-i.

Acoustic signals were recorded using the developed acoustic control system, which is a functional component of the ACS, mounted on the case of an axial compressor in the bearing area No. 1, in the immediate vicinity of the blades of the 0th, first and second stages. These blades are the most loaded elements of the axial compressor. Vibration signals were recorded using a Metrix SA6200A broadband vibration driver (USA). The output signals from the acoustic control system and vibration transmission via a coaxial cable through the audio module were fed to a PC with the Audacity high-frequency signal processing software package. The signal characteristics were as follows: sampling frequency – 44.1 kHz, resolution – 16 bits, amplitude corresponding to 32768 LSB ADC: +4 dBu (1.228 Vrms, 1.736 Vpk, 3.472 Vpp). In general, during the four years of operation of GPU type GTK-25-i, 36.878 experimental samples of signals were received for three of its technical states, which were stipulated by different periods of operation of GPU type GTK-25-i. So, "nominal" corresponds to the GPU state after repair, "defective" – the state of the decommissioned GPU (operating time to the established service life – 16 thousand hours) for repair work. The GPU with an operating time of 12 thousand hours was attributed to the "current" state. Of the total number of signal samples, 10,339 were received for the "nominal" state, 25,033 samples for the "current" state, and 1,506 samples for the "defective" state. The entire available dataset was divided into training, validation and test datasets in a ratio of 70:10:20.

#### 4. 2. Descriptive statistical analysis

Descriptive statistical analysis of signals was carried out using well-known approaches. Mathematical expectations, variances, standard deviations, minimum and maximum values, amplitude range and quartiles of signals were investigated and systematized. For example, Fig. 1 shows a histogram for the sample "defective" state of GPU type GTK-25-i on which the curve of the normal distribution is plotted, and Fig. 2 shows the Quantile-Quantile Plot (Q-Q Plot) of the acoustic signal of the "defective" GPU type GTK-25-i. A Q-Q plot is a scatter plot used in mathematical statistics as a graphical method for comparing two probability distributions by placing their quantiles against each other [14]. All graphical dependencies are built using the matplotlib library [15]. The study of the probability distribution of amplitude values was carried out using the Shapiro-Wilk test [16]. An example of a Box Plot [17] of acoustic signal amplitudes is shown in Fig. 3.

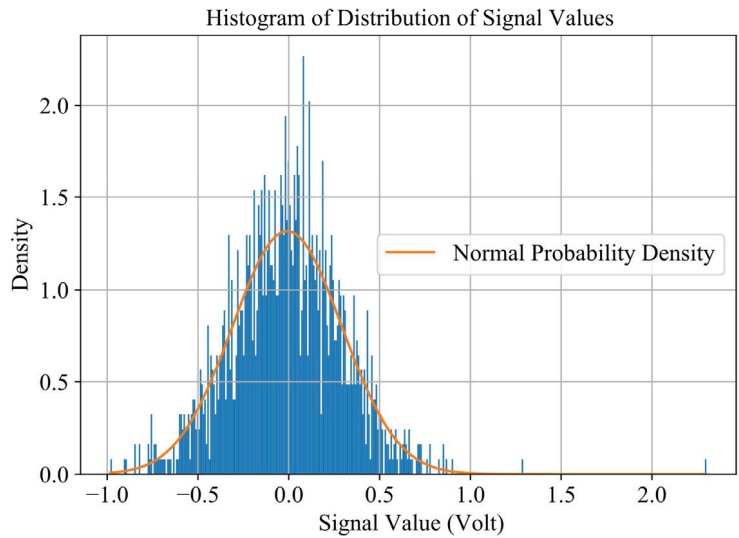

Fig. 1. Histogram of distribution of acoustic signal amplitudes for the "defective" state of the GPU type GTK-25-i

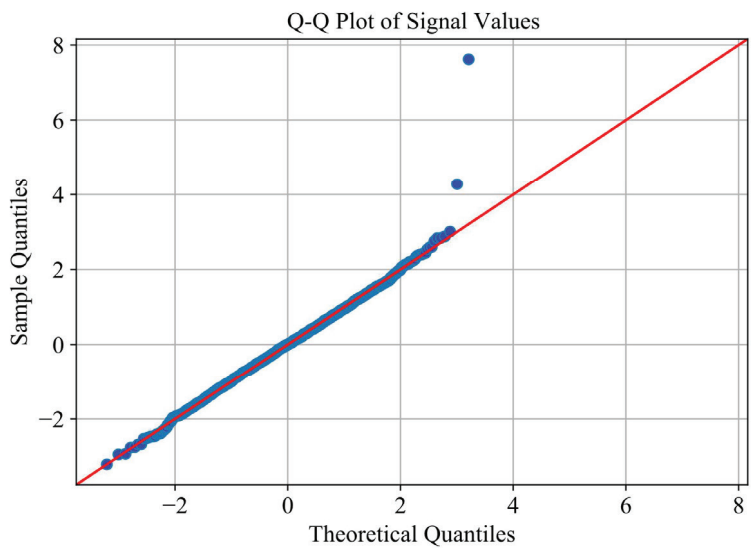

Fig. 2. Q-Q plot of normal distribution of the acoustic signal of the "defective" GPU type GTK-25-i

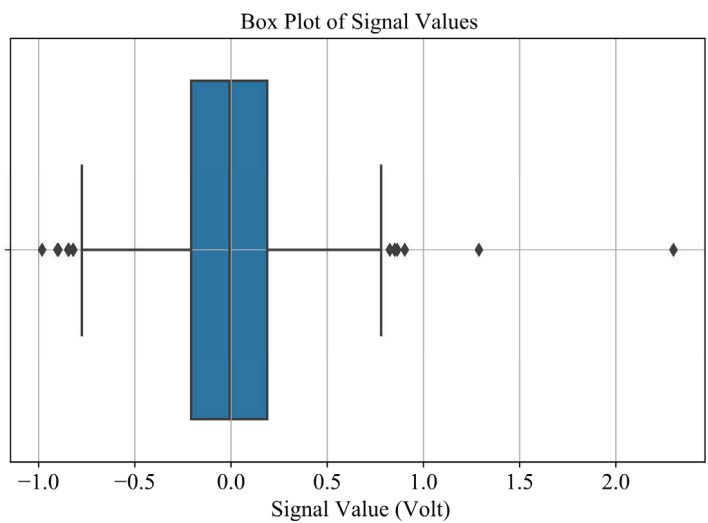

Fig. 3. Box plot of acoustic signal amplitudes of the "defective" GPU type GTK-25-i

It was found that there is no enough significant statistical evidence to reject the null hypothesis of the normal distribution of signal values (for the "nominal" state, the p-value is 0.859, for the "current" state, 0.093, and for the "defective" state, 0.071, at a significance level of 0.05).

So, further research and interpretation are carried out on the basis of the normal distribution of signal values.

### 4. 3. Time domain signal properties

The study of time properties was carried out using the visualization of time domain waveforms. An example of such visualization for the waveform of the acoustic signal of the "defective" state is shown in Fig. 4.

The study of autocorrelation properties was carried out in accordance with (1). An example of visualization is shown in Fig. 5.

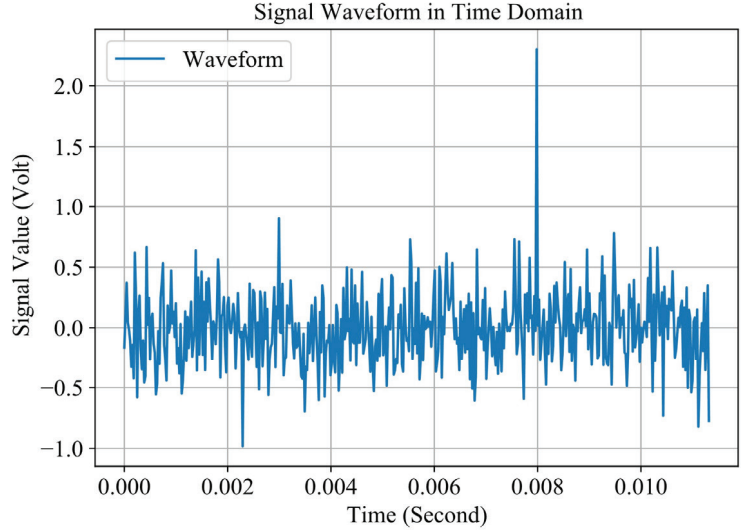

Fig. 4. Time domain waveform of the acoustic signal for the "defective" state of the GPU type GTK-25-i

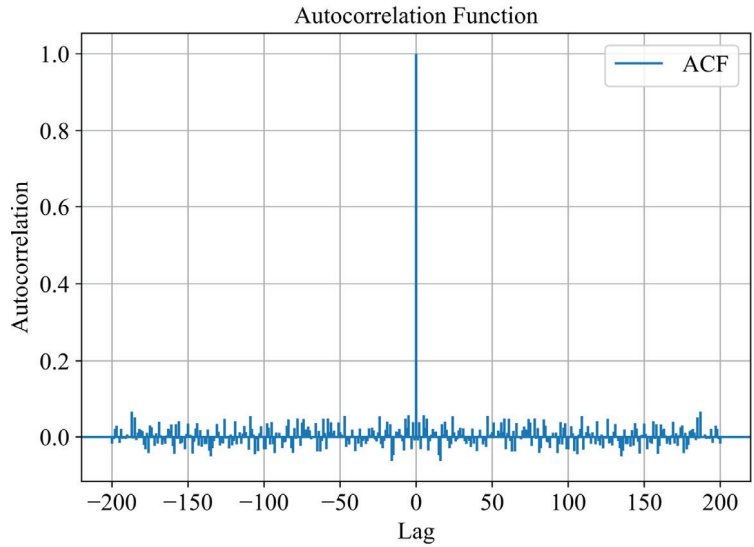

Fig. 5. Autocorrelation function of the acoustic signal for the "defective" state of the GPU type GTK-25-i

$$acf(l) = \frac{1}{N \cdot \mathrm{var}(x)} \cdot \sum_{n=0}^{N-1-l} (x_{n+l} - \bar{x}) \cdot (x_n - \bar{x}), \quad (1)$$

where $acf(l)$ – autocorrelation function (ACF) estimate; $x_n$ – sampled signal value, V; $N$ – total number of signal samples; $\mathrm{var}(x)$ – variance of signal $x$, V^2; $\bar{x}$ – sample mean, V; $l$ – lag, $l<(N-1)$; $n$ – time index, $n=0, 1,..., N-1$.

As can be seen from Fig. 4, 5, the signals are random with a rapidly decreasing autocorrelation function.

### 4. 4. Frequency domain properties

The energy spectral density (ESD) of acoustic and vibration signals was obtained using the Fast Fourier Transform (FFT) [18]. The FFT was calculated using the Python module numpy.fft according to (2).

$$S_x(k \cdot F) = T_s \cdot \sum_{n=0}^{N-1} x(n \cdot T_s) \exp(-j2\pi nk/N), \quad (2)$$

where $S_x(kF)$ – amplitude density, V/Hz; $x(nT_s)$ – sampled signal values, V; $N$ – total number of signal samples; $T_s$ – sampling interval in time, $T_s=1/f_s$, sec; $f_s$ – sampling frequency, Hz; $k$ – frequency index, $k=0, 1,..., N-1$; $n$ – time index, $n=0, 1,..., N-1$.

The energy spectral density was calculated at frequencies $kF$ as $|S_x(kF)|^2$ at intervals $F=1/NT_s$ in the frequency domain. An example for the acoustic signal of the GPU type GTK-25-i "defective" state is shown in Fig. 6.

An input data pipeline was developed, it provides the necessary transformation and aggregation of data, generation of sample batches. The model output data are categorical, therefore the target classes (GPU states) are encoded using the "one hot" encoding. Conversion to "one hot" is done in the input pipeline. Also, the generation of batches of features fed to the ANN input is carried out. Each feature sample is a combination of five maximum amplitude components of the acoustic signal and the standard deviation of the acoustic signal and five samples of the vibration signal and its standard deviation. The standard deviations are calculated directly in the input data pipeline in real time. All input signals coming to the input of the neural network are numerical in nature, their values are normally distributed. The pipeline is built with the tensorflow.data module.

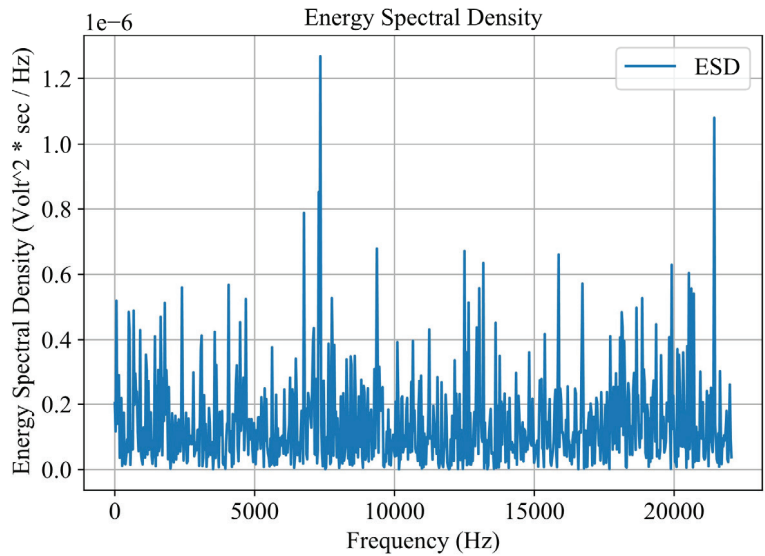

Fig. 6. Energy spectral density of the acoustic signal for the "defective" state GPU type GTK-25-i

### 4. 5. Neural network depth model design

The architecture of a deep fully connected feedforward neural network, trained by the backpropagation algorithm, was developed. Input ANN dimension is 12. The output layer contains 3 neurons for each state, respectively. ANN contains two hidden layers containing 256 and 128 neurons, respectively. The number of links between the input layer and the first hidden layer is 3,072, between the first and second hidden layers – 32,768, between the second hidden and output layers – 384 links. In total, the model has 38,195 parameters, of which 37,403 parameters are subject to change during training and 792 are constant parameters. This number of neurons and connections was chosen based on the need to provide the required network capacity. The hidden layer uses the ReLU [19] activation function, the output layer uses the SoftMax activation function. Batch normalization between the layers is applied.

The ANN was constructed using the frameworks TensorFlow [20], Keras [21], NumPy [22], pandas [23] in the Python 3 programming language [24] based on well-known approaches to digital signal processing using ANNs [25, 26]. The structure of neural connections is shown in Fig. 7, the structure of neuron layers and the dimensions of the data are shown in Fig. 8. The "Adam" [27] was used as an optimizer. The categorical cross-entropy (3) was used as the loss function; the efficient calculation of the entropy was carried out in accordance with [28]. The quality metric is overall accuracy (4) [29].

$$H(p,q) = -\sum_{x \in X} p(x) \cdot \log(q(x)), \quad (3)$$

where $H(p, q)$ – cross-entropy of distributions $p, q$; $p, q$ – probability distributions over the probability space $X$.

$$accuracy = \frac{TP + TN}{TP + TN + FP + FN}, \quad (4)$$

where $TP$ – number of true positive cases of classification, $TN$ – number of true negatives, $FP$ – number of false positives, $FN$ – number of false negative cases.

The Tensorboard web application was used to visualize training and validation scalars and structures of the neural network components.

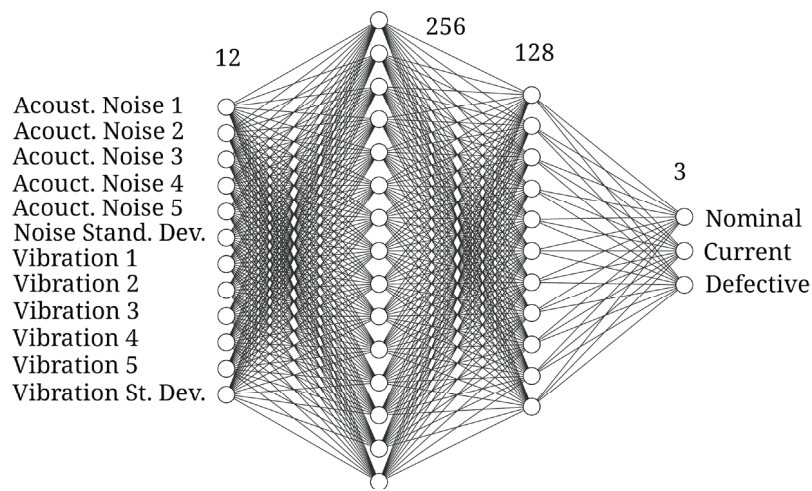

Fig. 7. Structure of neural connections

Industry control systems

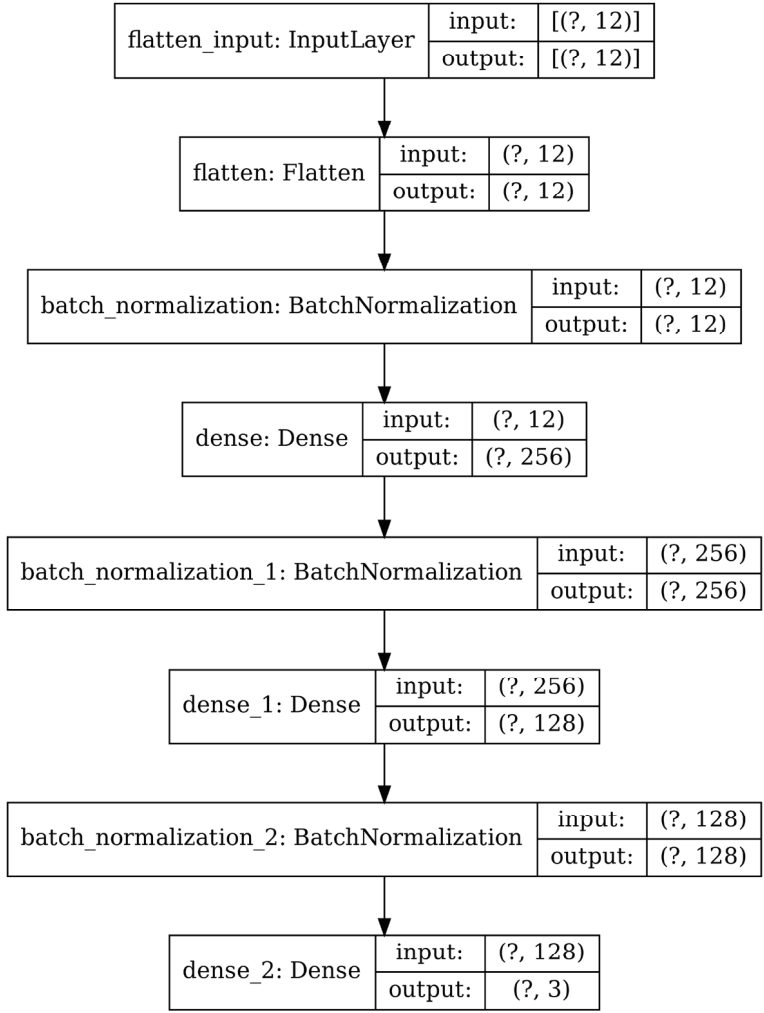

Fig. 8. Artificial neural network layer structure and data dimensions

### 4. 6. Training and validation

The model was trained on a specialized server equipped with a high-performance Tesla P4 GPU graphics adapter with 7611 MiB of video memory (NVIDIA-SMI 460.56, Driver Version: 460.32.03, CUDA Version: 11.2) manufactured by Nvidia (USA). The size of the training set is 5,309 signal samples, the size of the validation set is 590 samples. The training procedure lasts approximately 62.5 µs per sample, 2 ms per step (batch), 370 ms per epoch (including validation). The number of samples per one update of the optimizer gradient (batch size) is 32. During training, the training history is recorded, from which the indicators and the ability to visualize graphs characterizing the training quality were obtained. In particular, graphs showing the behavior of the loss function and accuracy metrics on the training and validation sets, respectively, are shown in Fig. 9. The assessment of these parameters is carried out in the training process, respectively, on the training and validation set in order to identify trends to overfitting or insufficient network capacity for generalization [30]. The values are calculated at the end of each epoch.

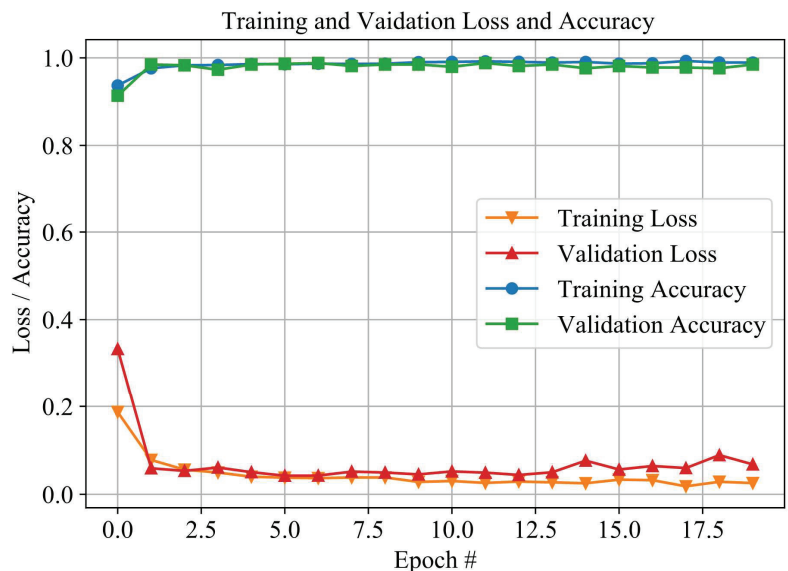

Fig. 9. Artificial neural network learning history graphs

The best system performance is obtained after 20 epochs, when the specified error level for the validation data is reached. As you can see, training accuracy reaches a value of 0.9904, validation accuracy reaches 0.9746. A further increase in the number of epochs does not lead to an improvement in quality indicators.

## 5. Results of studies of changes in the technical state of a gas pumping unit based on an artificial neural network

### 5. 1. Results of statistical, frequency and time series analysis

Descriptive statistical analysis of acoustic and vibration signals of gas pumping units of type GTK-25-i was done. The estimates of the expected values, variance, standard deviations, minimum and maximum values, the range of amplitude values and quartiles of signals were investigated and systematized (Table 1). Statistical estimates were obtained using the scipy.stats module [31], which has built-in functions for calculating many of the characteristics. The results were calculated for the entire available dataset separately for each of the states: "nominal", "current", and "defective".

Table 1

Statistical characteristics of signals, Volt

| state | "nominal" | | "current" | | "defective" | |
|---|---|---|---|---|---|---|
| statistics | noise | vibration | noise | vibration | noise | vibration |
| mean | –0.0002 | –0.0002 | –0.0006 | –0.0015 | –0.0093 | –0.0064 |
| std | 0.0247 | 0.0202 | 0.1601 | 0.2209 | 0.303 | 0.5514 |
| min | –0.0966 | –0.0704 | –0.6952 | –0.8825 | –0.9822 | –1.6656 |
| 25 % | –0.0167 | –0.0135 | –0.1051 | –0.1523 | –0.2073 | –0.3638 |
| 50 % | –0.0001 | 0.0002 | 0.0031 | –0.0007 | –0.0077 | –0.0056 |
| 75 % | 0.0162 | 0.0138 | 0.1087 | 0.1505 | 0.188 | 0.3284 |
| max | 0.094 | 0.0594 | 0.7219 | 0.7928 | 2.2995 | 8.4175 |

As you can see from Table 1, the statistical properties for "defective" and "nominal" conditions are significantly different. At the same time, the difference between the properties of the "current" and "defective" states is not significant. This is conditioned by its operating time – 12,000 hours, which approaches the established service life – 16,000 hours, which corresponds to the "defective" state of the GPU.

The probability distributions of signals were investigated, their normal nature was found (for the "nominal" state, the p-value is 0.859, for the "current" state, 0.093, and for the "defective" state, 0.071).

The study of the time series and autocorrelation properties was carried out. The frequency properties of acoustic and vibration signals were obtained. The possibility of using a combination of acoustic and vibration signals as input data of a neural network was found.

### 5. 2. Results of architecture design and neural network training

The architecture of a deep fully connected feedforward neural network, trained by the backpropagation algorithm, was developed. The input dimension is 12, the network contains two hidden layers with 256 and 128 neurons, respectively. The output layer contains 3 neurons for each state of the GPU, respectively. The training procedure lasts approximately 62.5 µs per sample, 2 ms per step (batch), 370 ms per epoch. The number of samples per update of the optimizer gradient (batch size) is 32. The "Adam" was used as the optimizer. Overall accuracy is a metric of quality. The overall accuracy was achieved at the level of 0.9904 on the training dataset and 0.9746 on the validation dataset.

### 5. 3. Results of testing the neural network model

The neural network was trained and tested on experimental data. The developed network was tested on the basis of a test set, the size of which is 1,475 signal samples. The value of test accuracy was obtained, which is 0.9871. The testing was done in a post-predict manner, when the test data go through the predict procedure. After that, the results of the neural network prediction were compared with the ground truth and a confusion matrix was obtained, presented in Table 2. The following class macro- and micro-averaged and weighted metrics were obtained from the confusion matrix: accuracy, true positive rate (TPR, recall), positive predictive value (PPV, precision), F1 metric, and the like. A detailed report with the quality indicators of the classifier is given in Table 3.

Table 2

Confusion matrix

| Ground Truth | Neural network prediction | | | |
|---|---|---|---|---|
| | "nominal" | "current" | "defective" | All states |
| "nominal" | 410 | 0 | 0 | 410 |
| "current" | 0 | 1,006 | 4 | 1,010 |
| "defective" | 0 | 15 | 40 | 55 |
| All states | 410 | 1,021 | 44 | 1,475 |

Table 3

Classification report

| State | Metric | | | |
|---|---|---|---|---|
| | precision | recall | f1-score | support |
| "nominal" | 1.0000 | 1.0000 | 1.0000 | 410 |
| "current" | 0.9853 | 0.9960 | 0.9906 | 1,010 |
| "defective" | 0.9091 | 0.7273 | 0.8081 | 55 |
| micro avg | 0.9871 | 0.9871 | 0.9871 | 1,475 |
| macro avg | 0.9648 | 0.9078 | 0.9329 | 1,475 |
| weighted avg | 0.9866 | 0.9871 | 0.9864 | 1,475 |
| samples avg | 0.9871 | 0.9871 | 0.9871 | 1,475 |

As you can see from Table 2, it was found during testing that for the "nominal" class, all 410 signal samples were recognized correctly (recall=1.0000). There are no cases of incorrectly recognized other classes as "nominal" (precision=1.0000). For the class "current", 4 of 1,010 samples were incorrectly qualified as other classes (recall=0.9960), also 15 samples of other classes were incorrectly qualified as "current" (precision=0.9853). For the class "defective", 15 of 55 samples were incorrectly classified as other classes (recall=0.7273) and 4 samples of other classes were mistakenly recognized as "defective" (precision=0.9091). Such quality metrics are quite acceptable and satisfy the requirements for the quality of diagnostics.

## 6. Discussion of the results of the study of the quality of GPU diagnostics based on a feedforward artificial neural network

Analysis of the frequency domain properties of acoustic noise and vibration signals (Fig. 6) shows that the signal spectrum is fairly uniform over a wide frequency range. This significantly complicates or makes it impossible to diagnose GPU based only on the spectral composition of the signals. The fast-decreasing autocorrelation function (Fig. 5) shows significant randomness of the signal values. With this nature of autocorrelation properties, it is practically impossible to apply well-known and developed techniques for analyzing and forecasting time series and making decisions based on them. An important result of the analysis of statistical properties (Fig. 1, 2) is the conclusion about the normal distribution of signal values. Normally distributed signals at a given energy level have the highest entropy value, therefore they are highly informative. Previous experience, including the experience of the authors, shows that using neural networks makes it possible to efficiently process such signals. The combination of acoustic noise and vibration signals increases the available amount of useful information entering the neural network and improves quality indicators compared to using only one of the listed parameters.

The main results of the study show that the use of simple feedforward neural networks has acceptable quality characteristics. The architecture of the neural network proposed in this study (Fig. 7, 8) turned out to be effective for solving the problem. At the same time, its computational complexity is negligible. Such a neural network can be implemented and scalable both in software and hardware. It is important that the proposed method uses only two information parameters (acoustic noise and vibration), which greatly simplifies data preprocessing and reduces the complexity of the model. In other approaches, as a rule, a much larger number of parameters are used, for example, 14 in [4] and 15 parameters in [12]. The obtained computational complexity allows carrying out diagnostics continuously and in real time. A key feature of the proposed architecture is the application of batch normalization between all layers of the neural network. This speeds up data processing and improves network stability. It also makes it possible to accelerate the convergence of the model, make the training of layers more independent of other layers, and reduce the sensitivity to the initial weights values by eliminating the internal covariant shift [32].

The closest to the proposed approach are the approaches described in [5] and [9]. They are based on the use of classical probabilistic neural network architectures, as well as on advanced learning algorithms, in particular, a combination of Nguyen-Widrow and Levenberg-Marquardt learning algorithms. Comparison of the accuracy of the three closest technical solutions is presented in Table 4 (according to Table 3 and [5, 9]).

Table 4

Comparison of results

| Quality metric | Model architecture | | | |
|---|---|---|---|---|
| | Classical convolutional neural network | Classical probabilistic neural network | Combination of Nguyen-Widrow and Levenberg-Marquardt learning algorithms | Proposed approach (combination of acoustic and vibration signals, fully connected network) |
| General accuracy | 0.9572 | 0.95 | 0.98 | 0.9871 |

As you can see from Table 4, the gain of the proposed approach in comparison with the classical ones ranges from 2.99 % to 3.71 %. Compared to the well-known improved learning algorithm based on a combination of the Nguyen-Widrow and Levenberg-Marquardt methods, the accuracy gain is at least 0.71 %. An important difference between the results obtained is that for the study, real data were used, captured during the GPU operation, in contrast to other studies that use a synthesis of artificial data or augmentation, modeling in computational experiments, simulation experiments, or data from laboratory physical models of the GPU manufactured in scale down.

The results obtained should be considered useful for the practical application of the developed neural network for diagnosing the technical state of GPU with a gas turbine drive of various types and powers. The advantage of the proposed solution is the ability to quickly implement computational structures based on modern microcontrollers and single-board industrial computers and software frameworks, since they are easily scalable. Another advantage is the adaptability to changing conditions, environmental noise, and the like. This study has the limitation that only fully connected feedforward networks were studied. There is a need to analyze the possibility of using recurrent architectures, in particular, with a long short-term memory (LSTM), which may be the subject of further research.

## 7. Conclusions

1. Using the improved ACS of the GPU type GTK-25-I, arrays of acoustic and vibration signals for three of its classes – "nominal", "current" and "defective" with a total of 36,878 samples were obtained and their descriptive statistical analysis was carried out. It was found that acoustic and vibration signals have a normal distribution (p-value in the range from 0.071 to 0.859). The time domain, autocorrelation and frequency domain properties of signals were investigated. It was found that the signals are random in nature with a rapidly decreasing autocorrelation function and a wide frequency spectrum. In terms of their characteristics, such signals can be used as input features for a neural network model. For this model, it is proposed to use a combination of acoustic noise and vibration signals. An input data pipeline was built for batch feeding of data to the model.

2. The architecture of a deep fully connected feedforward neural network with an input dimension of 12 was developed, containing two hidden layers with 256 and 128 neurons, respectively, and an output layer with 3 neurons with an acceptable training time of no more than 370 ms per epoch. The overall accuracy is 0.9904 on the training dataset and 0.9746 on the validation dataset.

3. The developed model was tested on an independent test data set. The indicators of the state classification quality that are acceptable for practical application were obtained. The overall accuracy is 0.9871, the weighted average of the F1 metric is 0.9864. Such indicators make it possible to apply the model to classify the technical conditions of the GPU with sufficient accuracy. This will prevent the occurrence of GPU failures and increase their reliability. The developed model can be used for diagnostics of GPU of any type and power, easily adapts to changing parameters of the acoustic environment over a wide range.